%% file: paper.tex
\documentclass[conference,final]{IEEEtran}

\input setbmp

\input seteps
\usepackage{enumerate}
\usepackage{amsthm}
\usepackage{amscd}
\usepackage{rotating}
\usepackage{subfigure}
\usepackage{array}
\usepackage[noadjust]{cite}
\usepackage{caption2}
\usepackage{times,graphicx}
\usepackage{epsfig, subfigure}  
\usepackage{multirow}
\usepackage{amsmath}
\usepackage{amssymb}
\usepackage{mathtools}
\usepackage{bm}

\DeclareMathOperator*{\argmax}{argmax}

\begin{document}

\bstctlcite{IEEEexample:BSTcontrol}
%

\title{Improve Lexicon-based Word Embeddings By Word Sense Disambiguation}

\author{\IEEEauthorblockN{Yuanzhi Ke}
\IEEEauthorblockA{Department of Information and Computer Science\\
	Faculty of Science and Engineering, Keio University\\
	Keio University\\
	Yokohama, Japan\\
	Email: enshi@soft.ics.keio.ac.jp}
\and
\IEEEauthorblockN{Masafumi Hagiwara}
\IEEEauthorblockA{Department of Information and Computer Science\\
	Faculty of Science and Engineering, Keio University\\
	Keio University\\
	Yokohama, Japan\\
	Email: hagiwara@soft.ics.keio.ac.jp
}}

\maketitle

\begin{abstract}
There have been some works that learn a lexicon together with the corpus to improve the word embeddings. However, they either model the lexicon separately but update the neural networks for both the corpus and the lexicon by the same likelihood, or minimize the distance between all of the synonym pairs in the lexicon. Such methods do not consider the relatedness and difference of the corpus and the lexicon, and may not be the best optimized. 
In this paper, we propose a novel method that considers the relatedness and difference of the corpus and the lexicon. It trains word embeddings by learning the corpus to predicate a word and its corresponding synonym under the context at the same time. For polysemous words, we use a word sense disambiguation filter to eliminate the synonyms that have different meanings for the context. 
To evaluate the proposed method, we compare the performance of the word embeddings trained by our proposed model, the control groups without the filter or the lexicon, and the prior works in the word similarity tasks and text classification task. The experimental results show that the proposed model provides better embeddings for polysemous words and improves the performance for text classification.
\end{abstract}

\IEEEpeerreviewmaketitle

\section{Introduction}

Some prior works show that using lexicons to refine vector-space representations of words can improve the performance for word similarity estimation and topic estimation\cite{yu2014improving,xu2014rc,faruqui:2014:NIPS-DLRLW,Bollegala2016}.
These methods traverse the lexicon for all the words and their neighbors, try to maximize the conditional probability of the synonyms given every word, or minimize the distance of the embeddings of the synonym pairs.
However, one of the weaknesses is that the word embeddings trained in such way may not be the best optimized for either the corpus nor the lexicon.

In this paper, we propose a new method that concentrates on the intersection of the lexicon and the corpus. It trains word embeddings by learning to predicate a word and its synonym under the context at the same time. We use a word sense disambiguation filter to eliminate the synonyms that have different meanings under the context for polysemous words. We observed improvement over the prior works in our experiments including word similarity estimation, word analogy, and text classification. The experimental results show that the proposed method achieves better word representations for the tasks.

\section{The Proposed Model}
\label{sec:prop}

\begin{figure}[tb]
	\begin{center}
		\subfigure[The models in the prior works use the whole corpus and the whole lexicon.]{
			\includegraphics{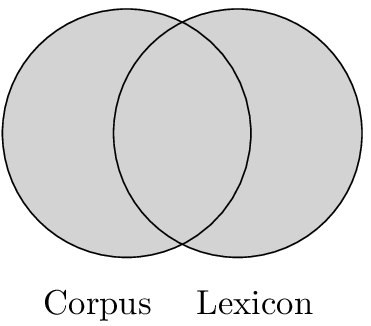}
			\label{fig:prior}
		}
		\hfill
		\subfigure[Our proposed model learns the corpus and the lexicon, and concentrates on the intersection of the corpus and the lexicon.]{
			\includegraphics{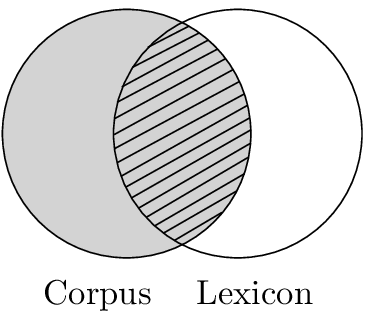}
			\label{fig:prop}
		}
	\end{center}
	\vspace{-0.3cm}
	\caption{Comparison of the objective of the prior works and the proposed model.}
	\label{fig:whattomodel}
\end{figure}

Because of the difference of the corpus and the lexicon, the word embeddings may not be the best optimized by simply combing the lexicon-based and corpus-based models or minimizing the distance for the embeddings of the synonym pairs. The difference may deteriorate both the corpus-based part and the lexicon-based part in such methods.

Thus, we explore another method that does not involve the relative complement of the corpus in the lexicon, and concentrates on the intersection as shown in Figure \ref{fig:prop}.
We let our proposed model predicate a word and its synonyms for the given context. For polysemous word, we employ a filter for the lexicon that chooses the correct paraphrases and eliminates the other for the current context. Therefore, only the synonyms of the corresponding senses are used to train the word embeddings.

\subsection{Objective} 

The objective is to maximize the joint conditional probability of a word and its paraphrases given the context words:

\begin{equation}
\label{eq:proposed_model}
\argmax_{V, \theta} \prod_{i}^{N} \left[P(w_i|C(w_i))\prod_{w_k \in R_i} f(w_i, w_k)P(w_k|C(w_i))\right].
\end{equation}

Here, $V$ is the matrix of all the word embeddings, $\theta$ is the hidden parameter matrix of the hidden layer, $w_i$ is the target word, $N$ is the size of the corpus, $C(w_i)$ is the context of $w_i$, $R_i$ is the paraphrase set of $w_i$, $w_k$ is one of the paraphrase. $f(x)$ is the filter function that eliminates the synonyms. 

We use a word sense disambiguation (WSD) filter. We compare the context and the gloss in WordNet \cite{Miller:1995:WLD:219717.219748,BA36267516} and choose the synonym that is the most likely to be the same in meaning by Lesk algorithm \cite{Lesk:1986:ASD:318723.318728}. Then the filter function returns $1$ for the chosen synonym, and $0$ for the others. It makes the model only learn the part of the lexicon that is related to the corpus.

\subsection{Training}

\begin{figure}[tb]
	\centering
	\epsfig{file=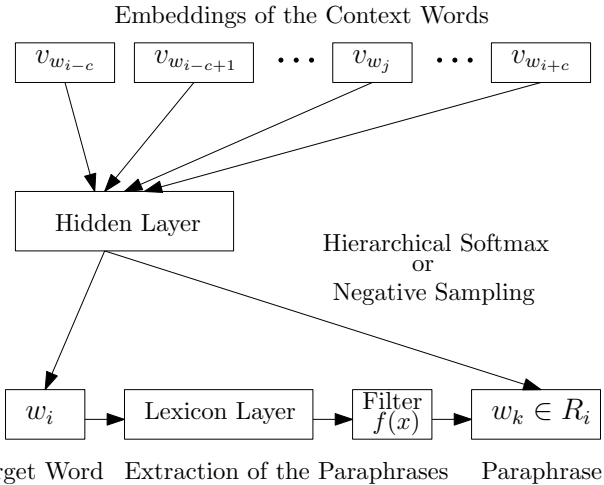, width=0.46\textwidth}
	\caption{The neural network to train the proposed model. $R_i$ refers to the paraphrase set of word $w_i$.}
	\label{fig:general}
\end{figure}

\ref{fig:general} shows the neural network to train the proposed model.

The log likelihood of Equation (\ref{eq:proposed_model}) is,
\begin{equation}
\label{eq:l1}
L = \sum_{i}^{N} \left[P(w_i|C(w_i))+\sum_{w_k \in R_i} f(w_i, w_k)P(w_k|C(w_i))\right].
\end{equation}

To maximize Equation (\ref{eq:l1}), approximately we maximize each $P(w_i|C(w_i))$ and $P(w_k|C(w_i))$. Let us denote the log likelihood of them as $L_{w_i}$ and $L_{w_k}$, respectively. As we use the same input words and the same layers, we mamximize $L_{w_i}$ and $L_{w_k}$ in the same way. We can train the model by hierarchical softmax or negative sampling similarly to word2vec.

\subsubsection{Training by Hierarchical Softmax}

For hierarchical softmax, at first we encode each word by a Huffman Tree \cite{huffman1952method}. Then let's denote $w$ as one of the target outputs of the neural network (can be $w_i$ or $w_k$), $\bm{x}_{w_i}$ as the averaged vector of the context words of word $w_i$, $l^w$ as the length of the code of $w$, $d^w_j$ as the $j$th code. Note that we use the context words of $w_i$ for both $w_i$ and $w_k$. We maximize $L_{w_i}$ and $L_{w_k}$ in the same way by $l^w$ step logistic regressions,

\begin{equation}
L_w = \log P(w|C(w_i)) = \log P(w|\bm{x}_{w_i}) = \sum_{j=1}^{l^w} L^j_w.
\end{equation}

$L^j_w$ is the log likelihood of the objective in the $j$th step. For the $j$th step, 

\begin{equation}
L^j_w = (1 - d^w_j) \log \left[\sigma (\bm{x}_{w_i}^\top\bm{\theta}^w_{j})\right] + d^w_j \log \left[1-\sigma (\bm{x}_{w_i}^\top\bm{\theta}^w_{j})\right].
\end{equation}

Here, $\bm{\theta} ^w_j$ is the hidden parameter vector for the $j$th logistic regression. 
The partial derivatives with respect to $\bm{\theta}^w_j$ and $\bm{x}_{w_i}$ are,

\begin{equation}
\frac{\partial L^j_w}{\partial \bm{\theta}^w_j} = \left[1-d^w_j-\sigma(\bm{x}_{w_i}^\top\bm{\theta}^w_{j})\right]\bm{x}_{w_i}.
\end{equation}
\begin{equation}
\frac{\partial L^j_w}{\partial \bm{x}_{w_i}} = \left[1-d^w_j-\sigma(\bm{x}_{w_i}^\top\bm{\theta}^w_{j})\right]\bm{\theta}^w_{j}.
\end{equation}

Then we update $\bm{\theta}^{w_i}_j$ by,
\begin{equation}
\bm{\theta}^{w_i}_{j}:=\bm{\theta}^{w_i}_{j}+\eta \left[1-d^{w_i}_j-\sigma(\bm{x}_{w_i}^\top\bm{\theta}^{w_i}_{j})\right]\bm{x}_{w_i},
\end{equation}

and update $\bm{\theta}^{w_k}_j$ by,

\begin{equation}
\bm{\theta}^{w_k}_{j}:=\bm{\theta}^{w_k}_{j}+\eta \left[1-d^{w_k}_j-\sigma(\bm{x}_{w_i}^\top\bm{\theta}^{w_k}_{j})\right]\bm{x}_{w_i},
\end{equation}

Then we update word embedding $\bm{v}_{w_c}$ for each context word $w_c$ by,
\begin{equation}
\begin{split}
\bm{v}_{w_c}:=&\bm{v}_{w_c}+\eta \sum^{l^{w_i}}_{j=1}\left[1-d^{w_i}_j-\sigma(\bm{x}_{w_i}^\top\bm{\theta}^{w_i}_{j})\right]\bm{\theta}^{w_i}_{j} + \\ 
&\eta \sum_{w_k \in R_i}f(w_i, w_k)\sum^{l^{w_k}}_{j=1}\left[1-d^{w_k}_j-\sigma(\bm{x}_{w_i}^\top\bm{\theta}^{w_k}_{j})\right]\bm{\theta}^{w_k}_{j}.
\end{split}
\end{equation}

\subsubsection{Training by Negative Sampling}

At first, we randomly draw noise words that are not equal to $w_i$ or $w_k$. Let's denote the set of such noise words as $N'(w_i, w_k)$. For each word in the corpus, we discriminate it from the noises.
Let's define $I_{w_i,w_k}^u$ for our model as,
\begin{equation}
I_{w_i,w_k}^u = \begin{cases}
1 & \text{ if } (u=w_i) \vee (u= w_k) ,\\ 
0 & \text{ if } (u\neq w_i) \wedge (u\neq w_k)  .
\end{cases}
\end{equation}

Let's denote $w$ as one of the target outputs of the neural network (can be $w_i$ or $w_k$),
\begin{equation}
L_{w} = \hspace{0.5cm}\sum_{\mathclap{u \in \{w_i, w_k\} \cup N'(w_i, w_k)}} \hspace{0.5cm} L_{w}^u .
\end{equation}
Here, $L_{w}^u$ is,
\begin{equation}
\begin{split}
L_{w}^u = &I_{w_i,w_k}^u \log \sigma (\bm{x}_{w_i}^\top \bm{\theta}^u) + \\
&(1-I_{w_i,w_k}^u) \log \left[1-\sigma (\bm{x}_{w_i}^\top \bm{\theta}^u)\right].
\end{split}
\end{equation}

Here, $\bm{\theta}^u$ is the hidden parameter vector for logistic regression to predicate if $u$ is equal to $w_i$ or $w_k$.  $\bm{x}_{w_i}$ is the average of the word embeddings of the context words of $w_i$. Note that we use the context words of $w_i$ for both $w_i$ and $w_k$. 

We can see that $L_{w_i}$ and $L_{w_k}$ are the same because $L_w^u$ only depends on $u$, $I_{w_i,w_k}^u$ and $\bm{x}_{w_i}$.

The partial derivatives of $L_w$ with respect to $\bm{\theta}^u$ and $\bm{x}_{w_i}$ are,

\begin{equation}
\frac{\partial L_{w}^u}{\partial \bm{\theta}^u} = \left[I_{w_i,w_k}^u-\sigma(\bm{x}_{w_i}^\top\bm{\theta}^u)\right]\bm{x}_{w_i}.
\end{equation}
\begin{equation}
\frac{\partial L_{w}^u}{\partial \bm{x}_{w_i}} = \left[I_{w_i,w_k}^u-\sigma(\bm{x}_{w_i}^\top\bm{\theta}^u)\right]\bm{\theta}^u.
\end{equation}

Then we update $\bm{\theta}^u$ for each $u \in \{w_i, w_k\} \cup N'(w_i, w_k)$ by,

\begin{equation}
\bm{\theta}^u:=\bm{\theta}^u+\eta \left[I_{w_i,w_k}^u-\sigma(\bm{x}_{w_i}^\top\bm{\theta}^u)\right]\bm{x}_{w_i}.
\end{equation}

Then we update word embedding $\bm{v}_{w_c}$ for each context word $w_c$ by,

\begin{equation}
\begin{split}
\bm{v}_{w_c}&:= \bm{v}_{w_c}+ \\
&\eta \hspace{0.1cm} \sum_{\mathclap{w_k \in \{w_i\} \cup R_i}} \hspace{0.5cm} f(w_i, w_k) \hspace{0.1cm} \sum_{\mathclap{u \in \{w_i, w_k\} \cup N'(w_i, w_k)}} \hspace{0.1cm} \left[I_{w_i,w_k}^u-\sigma(\bm{x}_{w_i}^\top\bm{\theta}^u)\right]\bm{\theta}^u.
\end{split}
\end{equation}

Here, we define $f(w_i, w_i) = 1$.


\section{Evaluation of the Proposed Model and the Filter Function}
\label{exp:filter}

\subsection{Experiment Setup}


To evaluate the effectiveness of the filter, we compared our proposed model with the control groups that use no filter function (models the union of the corpus and the lexicon) or no lexicon layer (i.e. CBOW model). 

For training, we used the first 100MB text data of wikipedia\footnote{http://mattmahoney.net/dc/text8.zip}. It contains 16,718,843 tokens.
At first we used the proposed model to train 50-dimensional word embeddings with different filter functions by negative sampling through 15 epochs, respectively. We set the context windows as 8. We let the models to draw 25 negative samples for negative sampling. The initial learning rate was set to 0.05.

We use intrinsic and extrinsic evaluation methods, including word similarity task and text classification task.

\subsection{Comparison of Word Embedding Spaces}

\begin{table*}[tb]
	\centering
	\caption{The closest words of frequent words ``have" and ``time", and rare words ``jig" and ``cobblers", in the vector spaces. Cosine similarities are used.}
	\begin{tabular}{|l|l|r|l|r|l|r|}
		\hline
		& \multicolumn{2}{l|}{No Lexicon (CBOW)}&\multicolumn{2}{l|}{No Filter} & \multicolumn{2}{l|}{WSD Filter (Proposed)} \\\hline
		{Word} & Close Word & Similarity & Close Word & Similarity & Close Word & Similarity \\\hline
		have & some  & 0.82  & some  & 0.83 &  some  & 0.82   \\
		& been  & 0.79  & although & 0.78 & although & 0.79   \\
		& however & 0.77  & many  & 0.77 & even  & 0.78   \\
		& many  & 0.77  & however & 0.77  & though & 0.78  \\
		& even  & 0.76  & been  & 0.77 & however & 0.78  \\
		& although & 0.76 & though & 0.77 & many  & 0.77   \\\hline
		time & before & 0.70  & before & 0.70 & next  & 0.72   \\
		& once  & 0.66  & next  & 0.70  & before & 0.70  \\
		& next  & 0.65  & when  & 0.63  & once  & 0.65  \\
		& when  & 0.64  & once  & 0.63  & when  & 0.64  \\
		& decade & 0.63  & again & 0.63 & again & 0.64   \\\hline
		jig  & polka & 0.74 & polka & 0.75 & merengue & 0.67  \\
		& aching & 0.65 & wha   & 0.68  & tits  & 0.66  \\
		& softies & 0.65 & rumba & 0.66 & rumba & 0.66   \\
		& earthy & 0.64 & supergroup & 0.66 & mambo & 0.64    \\
		& mellow & 0.64 & amour & 0.65 & piau  & 0.64   \\\hline
		cobblers & turnip & 0.61  & spoons & 0.77 & leather & 0.77   \\
		& jelly & 0.60  & seasoning & 0.72 & tanning & 0.75   \\
		& thorns & 0.59  & nori  & 0.72 & nori  & 0.74  \\
		& delicious & 0.59  & rawhide & 0.71  & leathers & 0.71  \\
		& cuttlefish & 0.58  & necklaces & 0.70  & feta  & 0.71  \\\hline	\end{tabular}%
	\label{tab:sample_word}%
\end{table*}%

To see how our proposed model makes a difference, we compared the closest words for some frequent and rare words based on the word embeddings trained in the experiment.  \ref{tab:sample_word} shows the top five closest words of two frequent content words in text8, ``have" and ``time"; and those of two rare content words, ``jig" and ``cobblers" as an example. ``jig" is a kind of dance. ``cobblers" can mean the craftsman that repairs shoes, or a kind of dessert or cocktail. 

For frequent words, the closest words in each group are almost the same. 
However, for the rare words whose frequencies are low, the proposed model makes more differences. For the rare word example ``jig" that means a kind of dance, let us count the number of words related to dance in the top five closest words in each group. In ``No Lexicon" group, there is one word related to dance (``polka") among the top 5 closest words of ``jig". In ``No Filter" group, there are two words related to dance (``polka" and ``rumba"). In ``WSD Filter" group, there are three words related to dance (``merengue", ``rumba", ``mambo"). The proposed model with WSD filter captures the meaning of ``jig" better.

For the rare word example ``cobblers", let us denote the meaning of ``craftsman that repairs shoes" as $sense_{shoe}$, the meaning of ``a kind of dessert or cocktail" as $sense_{food}$. In the top five closest words in ``No Lexicon" group, there are four words related to $sense_{food}$ (``turnip", ``jelly", ``delicious", ``cuttlefish"), no one is related to $sense_{shoe}$. In the top five closest words in ``No Filter" group, there are three words related to $sense_{food}$ (``spoons", ``seasoning", ``nori"), one related to $sense_{shoe}$ (``rawhide"). In the top five closest words in ``WSD Filter" group, there are two words related to $sense_{food}$ (``nori", ``feta"), three words related to $sense_{shoe}$ (``leather", ``leathers", ``tanning"). We can see that the word embedding of ``cobblers" trained in ``No Lexicon" group is almost entirely fitted to $sense_{food}$, the embedding in ``No Filter" group is close to $sense_{food}$ as well but also fitted to $sense_{shoe}$, the embedding trained by WSD filter is the most balanced for both $sense_{food}$ and $sense_{shoe}$.

We can see that learning the lexicon together with the corpus can help the word embeddings learn secondary meanings for rare words, and the WSD filter improves it further more and helps polysemous words be close to both their primary related words and secondary related words, avoiding overfitting to any one of them.

\begin{table*}[tb]
	\centering
	\caption{Comparison of the Spearman's rank correlations $\rho \times 100$ with human assigned similarity in WordSim353 dataset.}
	\begin{tabular}{|l|r|r|r|}\hline
		& WordSim353\cite{2002:PSC:503104.503110} & Polysemous WordSim353 & SCWS\cite{huang2012improving} \\\hline
		No Lexicon (CBOW) & 68.44 & 68.54 & 63.49 \\\hline
		No Filter & 68.75 & 68.89 & 63.01 \\\hline
		WSD Filter (Proposed) & \textbf{69.54} & \textbf{69.93} & \textbf{63.87} \\\hline
	\end{tabular}%
	\label{tab:filters}%
\end{table*}%

\subsection{Comparison with Human Assigned Similarity}

We also evaluated the performance by comparing the correlations of human assigned similarities and those assigned by the word embbeddings trained in different groups. We used the following datasets:
\begin{itemize}
	\item WordSim353\cite{2002:PSC:503104.503110}: This dataset contains 353 word pairs. Each pair is annotated with similarity scores assigned by 13 human subjects. We use the averaged value of the scores. We extracted a list of polysemous words that have more than one synonyms in WordNet and found that there are 319 word pairs in WordSim353 that contain polysemous words by comparison with the list. 
	\item Polysemous WordSim353: To concentrate on the evaluation for polysemous words, we extracted the word pairs from WordSim353 that either one of it is in the polysemous word list extracted from WordNet. We got 319 word pairs. We used the same similarity scores in WordSim353 for this dataset.
	\item SCWS\cite{huang2012improving}: It contains 2,003 word pairs, with sentences containing the words. Each pair is annotated with similarity scores assigned by 10 human subjects. We used the averaged value. We also compared the word pairs and the polysemous list extracted from WordNet and found that 1,878 pairs contain at least one word in the list.
\end{itemize}

To evaluate the model, 
we annotated the word pairs with the cosine similarity scores of their word embeddings, then calculated the Spearman's rank correlation $\rho$ of the cosine similarity scores and the human assigned scores. 

The results are summarized in  \ref{tab:filters}. With the WSD filter, the proposed model achieves the best performance in the word similarity task. For the dataset that contains only polysemous words, the improvement by WSD filter is more significant.

We can see that the proposed model with WSD filter is closer to human in estimating the similarities of words, especially for polysemous words.

\subsection{Text Classification Tasks}

\begin{table}[htb]
	\centering
	\caption{Comparison of the performance of text classification for ``20 newsgroup dataset".}
	\begin{tabular}{|l|r|}
		\hline
		& \multicolumn{1}{c|}{Validation Accuracy \%} \\\hline
		No Lexicon (CBOW) & 66.59 \\\hline
		No Filter &  67.42 \\\hline
		WSD Filter (Proposed) & \textbf{69.49} \\\hline
	\end{tabular}%
	\label{tab:tc}%
\end{table}%

A common usage of word embeddings is text classification. To compare the performance of the word embeddings in this task, we used the ``20 newsgroup dataset"\footnote{http://www.cs.cmu.edu/afs/cs.cmu.edu/project/theo-20/www/data/news20.html}. The dataset contains 20,000 news from 20 different groups. The objective is to classify the news into their correct groups. 20\% of the dataset were randomly chosen for validation, and the other samples were used for training.

We used a public implement of a convolutional neural network that takes the pretrained word embeddings to classify texts\footnote{https://github.com/fchollet/keras/blob/master/examples/\\pretrained\_word\_embeddings.py}.
We used the 50-D word embeddings trained in the previous experiments as the input. 

The results are shown in \ref{tab:tc}. The models involving lexicons outperform CBOW model. The WSD filter of the proposed model improves the performance further more and we can observe a significant improvement for the validation set.

\subsection{Discussion}

From the experimental results, we can see that dropout of the relative complement of the corpus in the lexicon by the WSD filter improves the performance of the trained vectors when they are applied to estimate word similarity or classify texts. Especially, it helps the model learn the rare words and discover the other meanings. It makes the relationship between word embeddings more like the human judgment. Using the word embeddings trained by the proposed model with WSD filter, the convolutional neural network classifier can achieve better accuracy for news text.

\section{Comparison with the Prior Works}
\label{sec:comp}

\newcommand{\tabincell}[2]{\begin{tabular}{@{}#1@{}}#2\end{tabular}}  
\begin{table}[tb]
	\centering
	\caption{Comparison of the rank correlation score on WordSim353 with the other  works that uses lexicons to improve the single-prototype word embeddings on the basis of word2vec.}
	\begin{tabular}{|l|r|r|r|}
		\hline
		& \multirow{2}*{WordSim353} & \multicolumn{2}{c|}{Word Analogy} \\\cline{3-4}
		&  & Semantic & Syntactic \\\hline
		JRCM\cite{yu2014improving}  & 53.70 &   -   & 29.90  \\\hline
		R-net\cite{xu2014rc} &  - & 32.64  & 43.46\\\hline
		C-net\cite{xu2014rc} & 68.30 & 37.07  & 40.06 \\\hline
		RC-net\cite{xu2014rc} &  - & 34.36  & 44.42\\\hline
		\tabincell{l}{Retrofitting \\ \cite{faruqui:2014:NIPS-DLRLW}} & 58.40 &  -    & 52.50 \\\hline
		Proposed & \textbf{68.97} & \textbf{59.23}  & \textbf{57.19}  \\\hline
	\end{tabular}%
	\label{tab:comp353}%
\end{table}%

In this experiment, we compared our model with the other methods that use lexicons to learn word embeddings on the basis of word2vec including JRCM, RC-Net, and a Retrofitting method\cite{faruqui:2014:NIPS-DLRLW} that refine pre-trained word embeddings by a semantic lexicon. All of them use the whole lexicon.

We used the proposed model to train 300-dimensional vectors with enwiki9\footnote{http://mattmahoney.net/dc/enwiki9.zip}, set the context window as 5, the same with the previous work\cite{xu2014rc}. We let the model to draw 15 negative samples and trained the vectors through 4 epochs to ensure there was no underfitting. 

We used WordSim353 and Google's Word Analogy Task \cite{DBLP:journals/corr/abs-1301-3781} for comparison of the performance, which are used in the other works. 

The Google's Word Analogy Task dataset contains pairs of word pairs in similar relationships. The task is to predicate the related word for the given word whose relationship is similar to the given word pairs. For example, for two word pairs, ``France : Paris" and ``Japan : Tokyo", the task is to predicate ``Tokyo", given ``Japan" and ``France : Paris". We used the original method in \cite{DBLP:journals/corr/abs-1301-3781} for the tasks. For example, at first we get the vector $\bm{v}_{France}-\bm{v}_{Paris}+\bm{v}_{Tokyo}$. Then we find the word whose vector is the closest to it as the answer. The dataset can be divided into two parts, one part contains word pairs that is semantically related, while the other part contains the syntactically related ones.



In \ref{tab:comp353}, we compare our achieved rank correlation score on WordSim353 and achieved correct rate on Google's word analogy task with the reported scores of the other works. We can see that our proposed model outperforms the others. With WSD filter, our proposed model achieves the best performance in the two tasks.

\section{Related Works}
\label{sec:related}

Relation Constrained Model\cite{yu2014improving} is a model to learn word embeddings from a lexicon. It maximizes the probability of the related words of each word:

\begin{equation}
\frac{1}{N}\sum_{i=1}^{N}\sum_{w\in R_{w_i}} \log p(w|w_i).
\end{equation}

Here, $N$ is the total number of the vocabulary. $R_{w_i}$ is relation sets of word $w_i$.
The authors also join Relation Constrained Model to CBOW and call it Joint Relation constrained Model(JRCM), whose objective is as the following:

\begin{equation}
\frac{1}{T}\sum_{t=1}^{T}\log p(w|context(w))+\frac{C}{N}\sum_{i=1}^{N}\sum_{w\in R_{w_i}} \log p(w|w_i).
\end{equation}

Here, $T$ is the size of the corpus. $C$ is the weight to join relation constrained model. It shares the word embeddings for CBOW and RCM.

RC-net\cite{xu2014rc} is another model that joints semantic networks and word2vec. It minimizes the distance of the connected nodes in the semantic network and that of the words in the same category. The objective of RC-net\cite{xu2014rc} is to minimize the following,

\begin{equation}
\label{rnet}
\sum_{(\bm{h},\bm{r},\bm{t})\in R}\sum_{(\bm{h}',\bm{r},\bm{t}')\in R'}\left[\bm{r}+d(\bm{h}+\bm{r},\bm{t})-d(\bm{h}'+\bm{r},\bm{t}')\right]_+.
\end{equation}

and,

\begin{equation}
\label{cnet}
\sum_{i=1}^{V}\sum_{j=1}^{V}S(w_i,w_j)d(w_i, w_j).
\end{equation}

At the same time, they also minimize the objective of Skip-gram. That is,
\begin{equation}
\sum_{w\in N}^{|N|}\log p(context(w)|w).
\end{equation}

Equation (\ref{rnet}) is called R-net. It minimizes the distance from a word to the sum of its related words and the relationships. $\left[x\right]_+$ means the positive part of $x$. $(\bm{h},\bm{r},\bm{t})$ refers that the word embeddings $\bm{h}$ and $\bm{t}$ are related, the relationship vector is $\bm{r}$. $R$ is the set of relations. $R'$ is the noise set containing corrupt relations, defined as:
\begin{equation}
R'(\bm{h},\bm{r},\bm{t})=\left \{ (\bm{h}',\bm{r},\bm{t})|\bm{h}' \in W\right \} \cup \left \{(\bm{h},\bm{r},\bm{t}')|\bm{t}' \in W\right \}.
\end{equation}
Here, $W$ is the set of word embeddings.

Equation (\ref{cnet}) is called C-net. It minimizes the distance of words in the same category, but if the category is larger, its members are farther to each other. $V$ in Equation (\ref{cnet}) refers to a category. $w_i$ and $w_j$ are two of the members of $V$. $S(w_i,w_j)$ is the score of them, defined as:
\begin{equation}
\sum_{i}^{V} \sum_{j}^{V} S(w_i, w_j) = 1.
\end{equation}

\cite{faruqui:2014:NIPS-DLRLW} proposed a method to refine pretrained word embeddings. It is not limited to word2vec but can be used for the word embeddings trained by any model. For a semantic network whose collection of edges is $E$, it minimizes,
\begin{equation}
\sum_{N}^{i=1}\left [\alpha_i ||\bm{q_i}-\hat{\bm{q}}_i||^2 + \sum_{(i,j)\in E} \beta_{ij} ||\bm{q_i}-\hat{\bm{q}}_j||^2 \right].
\end{equation}

Here, $\hat{\bm{q}}$ is the pretrained vector, and $\bm{q}$ is the vector to output. This model minimizes the distance of related words.

We can see that in the prior works, the lexicon is traversed separately from corpus. They can be seen as models of the union of the corpus and the lexicon to train the word embeddings.

\section{Conclusion}
\label{sec:conclusion}



To employ lexicons to train quality word embeddings, we proposed a novel model that concentrates on the the intersection of the corpus and the lexicon, predicates a word and its synonyms for the given context. For polysemous words, to avoid the synonyms that are in the lexicon but not expressing the same meaning under the given context, we employ a WSD filter to choose the proper synonyms by Lesk algorithm. The experimental results indicate that the word embeddings are improved by such dropout and outperform those trained by the prior works that use the whole lexicon. They show that the dropout by the WSD filter helps the discovery of the meanings of the rare words, alleviates the overfitting to one of the meanings for polysemous words and improves generalization for the text classification task. 

\bibliographystyle{IEEEtran}
\bibliography{references}

\end{document}

%% file: setbmp.tex
\def\centerbmp#1#2#3{\vskip#2\relax\centerline{\hbox to#1{\special
  {bmp:#3 x=#1, y=#2}\hfil}}}

%% file: seteps.tex
\def\centereps#1#2#3{\vskip#2\relax\centerline{\hbox to#1{\special
  {eps:#3 x=#1, y=#2}\hfil}}}